\documentclass[lettersize,journal]{IEEEtran}
\usepackage{amsmath,amsfonts}
\usepackage{array}
\usepackage[caption=false,font=normalsize,labelfont=sf,textfont=sf]{subfig}
\usepackage{textcomp}
\usepackage{stfloats}
\usepackage{url}
\usepackage{verbatim}
\usepackage{graphicx}
\usepackage{cite}
\usepackage{multicol}
\usepackage[bookmarks=true]{hyperref}
\usepackage{amssymb}            
\usepackage{mathtools}          
\usepackage{mathrsfs}           
\usepackage{graphicx}
\usepackage{subfig}
\usepackage{subcaption}         
\usepackage[space]{grffile}     
\usepackage{url}                
\usepackage{lipsum}             
\usepackage{xcolor}
\newcommand{\red}[1]{\textcolor{red}{#1}}
\newcommand{\teal}[1]{\textcolor{teal}{#1}}
\usepackage{algorithm}
\usepackage{algpseudocode}
\usepackage{caption}
\usepackage{float}
\usepackage{amsmath}
\usepackage{wrapfig}  
\usepackage{afterpage}

\newcommand{\Lift}{\emph{Lift}\ }
\newcommand{\Can}{\emph{Can}\ }
\newcommand{\Square}{\emph{Square}\ }

\begin{document}

\title{Accelerating Residual Reinforcement Learning \\with Uncertainty Estimation}


\author{Lakshita Dodeja$^{1}$, Karl Schmeckpeper$^{2}$, Shivam Vats$^{1}$, Thomas Weng$^{2}$, \\ Mingxi Jia$^{1}$, George Konidaris$^{1}$, and Stefanie Tellex$^{1,2}$ %
\thanks{Manuscript received: June, 24, 2025; Accepted October, 22, 2025.}
\thanks{This paper was recommended for publication by Editor Jens Kober upon evaluation of the Associate Editor and Reviewers' comments.
This work was supported by NASA-Kennedy under award number 80NSSC23M0075} 
\thanks{$^{1}$Lakshita Dodeja, Shivam Vats, Mingxi Jia, and George Konidaris are with Brown University, Providence, RI, USA 
        {\tt\footnotesize lakshita\_dodeja@brown.edu}}%
\thanks{$^{2} $Karl Schmeckpeper and Thomas Weng are with the Robotics and AI Institute, Cambridge, MA, USA.}%
\thanks{$^{1,2} $Stefanie Tellex is with both Brown University and the Robotics and AI Institute.}%
\thanks{Digital Object Identifier (DOI): see top of this page.}}

\markboth{IEEE ROBOTICS AND AUTOMATION LETTERS. PREPRINT VERSION. ACCEPTED October, 2025}%
{Shell \MakeLowercase{\textit{et al.}}: A Sample Article Using IEEEtran.cls for IEEE Journals}

\IEEEpubid{0000--0000/00\$00.00~\copyright~2021 IEEE}

\maketitle
\begin{figure*}[t!]
    \centering
    \includegraphics[width=0.75\textwidth]{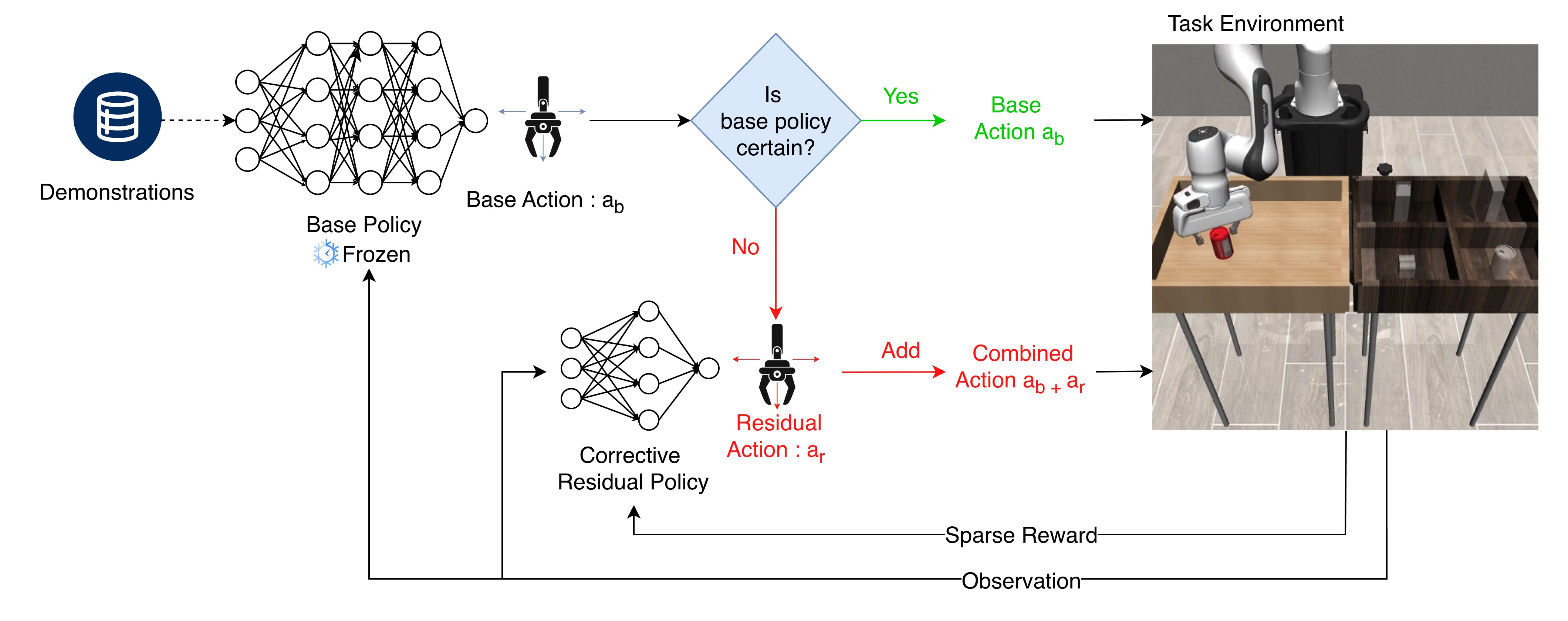}
    \caption{We propose two improvements to accelerate Residual RL : 1) We use uncertainty estimation to constrain exploration around the base policy. 2) We modify the off-policy critic to learn the $Q$ function for the combined action. We test our method with two uncertainty metrics, i.e. distance to data and ensemble variance.}
    \label{fig:title_figure}
\end{figure*}
\begin{abstract}
Residual Reinforcement Learning (RL) is a popular approach for adapting pretrained policies by learning a lightweight residual policy that provides corrective actions. While Residual RL is more sample-efficient than finetuning the entire base policy, existing methods struggle with sparse rewards and are designed for deterministic base policies.
We propose two improvements to Residual RL that further enhance its sample efficiency and make it suitable for stochastic base policies.
First, we leverage uncertainty estimates of the base policy to focus exploration on regions in which the base policy is not confident. Second, we propose a simple modification to off-policy residual learning that allows it to observe base actions and better handle stochastic base policies. We evaluate our method with both Gaussian-based and Diffusion-based stochastic base policies on tasks from Robosuite and D4RL, and compare against state-of-the-art finetuning methods, demo-augmented RL methods, and other Residual RL methods. Our algorithm significantly outperforms existing baselines in a variety of simulation benchmark environments. We also deploy our learned policies in the real world to demonstrate their robustness with zero-shot sim-to-real transfer. Paper homepage :  \href{http://lakshitadodeja.github.io/uncertainty-aware-residual-rl/}{lakshitadodeja.github.io/uncertainty-aware-residual-rl/}
\end{abstract}

\section{Introduction}

\label{sec:introduction}
\IEEEPARstart{R}{esidual} Reinforcement Learning improves the performance of pretrained policies by training a separate policy to output corrective actions ~\cite{silver2018residual, johannink2019residual}. Directly fine-tuning the pretrained policy is often computationally expensive, especially in the case of policies with a large number of parameters~\cite{chi2024diffusionpolicy, black2024pi_0}, and is prone to instability~\cite{mnih2016asynchronous}. In contrast, Residual RL provides an efficient alternative to refine the base policy with minimal additional computation. This residual correction enables the agent to make targeted improvements however the base policy was built.

Despite the promise of Residual RL, existing algorithms suffer from unconstrained exploration, often requiring extensive online interaction and dense reward shaping to achieve meaningful improvements~\cite{liu2022deep, zhang2024residual}. Furthermore, recent advancements in imitation learning leverage highly effective stochastic policies: Gaussian Mixture Model-based policies~\cite{mandlekar2021matters} and Diffusion policies~\cite{chi2024diffusionpolicy} excel at modeling complex, multi-modal distributions. 
In such cases, residual RL algorithms that assume a deterministic base policy~\cite{silver2018residual, johannink2019residual} are not suitable. 

We address these limitations by proposing two improvements to Residual RL that enhance its sample efficiency and make it more suitable for stochastic policies. First, we leverage uncertainty estimates from the base policy to guide the exploration of the residual policy. Our key insight is that regions where the base policy is confident require minimal exploration by the residual agent, allowing it to focus exploration on areas of high uncertainty. This targeted exploration significantly improves the sample efficiency of residual learning.

Second, existing off-policy Residual RL algorithms learn the $Q$ function only for the residual action $a_r$, i.e. $Q(s, a_r)$, implicitly assuming that the base policy's action can be inferred from the state $s$. This is insufficient when dealing with stochastic base policies, since they sample different actions given the same state. In the stochastic setting, the Residual RL agent cannot infer a single base action, making it difficult to learn a good residual action. Prior works have attempted to solve this by using learned bottleneck features of the base policy as a prior for residual learning~\cite{alakuijala2021residual} or by augmenting the observed state with the base action for on-policy learning~\cite{ankile2024imitation}. We propose an asymmetric actor-critic approach, in which the critic learns the $Q$ function for the fully observed action executed in the environment, comprising both the base action and the residual action, while the actor learns partial residual actions only. This formulation ensures that information about the stochastic base actions is available to the $Q$ function while also making the critic invariant to the split between the residual and base action.

\IEEEpubidadjcol

We evaluate our approach on a variety of manipulation tasks from Robosuite \cite{mandlekar2021matters} and Franka Kitchen tasks from D4RL \cite{fu2020d4rl}. We test our approach on both Gaussian Mixture Model and Diffusion base policies. Finally, we compare our approach with a state-of-the-art finetuning method \cite{ren2024diffusion}, a demo-augmented RL method \cite{hu2023imitation}, and other Residual RL methods \cite{yuan2024policy, silver2018residual, johannink2019residual}. Our proposed approach outperforms or is comparable to other baselines in all tasks. We also perform several ablation studies to test various aspects of our algorithm and deploy the learned policies on a real robot to demonstrate sim-to-real transfer. Our proposed approach is visualized in Figure~\ref{fig:title_figure} with the main contributions summarized as follows:

\begin{enumerate}
    \item We present a novel algorithm to accelerate Residual Reinforcement Learning using uncertainty estimates.
    \item We modify off-policy Residual Reinforcement Learning to work with stochastic base policies using an asymmetric actor-critic approach. 
    \item We validate our method on robotic manipulation tasks from different simulators and against several baselines. We demonstrate zero-shot sim-to-real transfer of our learned policies.
\end{enumerate} 

\section{Background}
\label{sec:problem_setup}
In Residual RL, we assume that we have a suboptimal base policy $\pi_{b}$. The objective is to learn a lightweight residual policy $\pi_{r}$, on top of the base policy that gives a corrective action $a_{r}$, producing a more accurate and robust combined policy. Thus, the final policy executed in the environment is: 
\begin{equation}
    \pi = \pi_b(s) + \pi_r(s)
\end{equation}
Residual RL transforms the original Markov decision process (MDP) formulation $M = <S, A, R, T, \gamma>$ to the residual MDP (RMDP) formulation $ M_{r}= <S, A_{r}, R, T_{r}, \gamma>$~\cite{silver2018residual}, where $S$ is the set of states, $A_{r}$ is the set of residual actions, $R$ is the reward received for taking action $a_{r}\in A_{r}$ in state $s\in S$, $T_{r}$ is the probability of taking action $a_r$ in state $s$ and ending up in a new state $s'$ and $\gamma$ is the discount factor. The residual transition function can be converted back to the original transition function as follows: 
\begin{equation}\label{eqn:transition}
    T_{r}(s, a_r, s') = T(s, \pi_{b}(s) + a_r, s').
\end{equation}

\section{Related Work}
\textbf{Residual RL for Stochastic Base Policies.} Residual RL, first introduced for robotics in~\cite{johannink2019residual} and~\cite{silver2018residual}, learns a corrective residual policy over a base controller, which can be either hand-designed or derived from model-predictive control. Importantly, these methods assume a deterministic base controller, as the residual policy is not conditioned on the base action. However, current state-of-the-art imitation learning algorithms like Diffusion policy \cite{chi2024diffusionpolicy} and GMM-based policies \cite{mandlekar2021matters} are non-deterministic, making the original Residual RL formulation insufficient due to the lack of information about the base policy. Some works introduce noise in the base action to enhance robustness and induce stochasticity \cite{liu2022deep, zhang2024residual}. Other works inform residual learning about the base policy by conditioning it on the learned bottleneck features of the base policy~\cite{alakuijala2021residual}, and incorporating the base action in the observed state to inform the residual policy for on-policy learning~\cite{ankile2024imitation}. In contrast, our work modifies off-policy RL to learn the $Q$ function for the combined action taken in the environment (i.e. the sum of base and residual action), and the actor uses the same $Q$ function to select a residual action. Therefore, our Residual RL formulation can handle the stochasticity of the base policies by making the base action observable to the critic while also being invariant to the split between the base action and residual action. Policy Decorator \cite{yuan2024policy} also uses the combined action as an input to their critic for Residual RL, but we do a thorough quantitative evaluation to show that it significantly improves performance. 

\textbf{Imitation Learning and Residual RL.} Several works have explored the integration of Imitation Learning (IL) base policies with Residual RL. Residual RL has been applied to insertion tasks~\cite{schoettler2020deep}, where demonstrations are incorporated as an auxiliary behavior cloning (BC) loss during RL training~\cite{nair2018overcoming}. FISH~\cite{haldar2023teach} employs a non-parametric base policy alongside a residual policy that uses optimal transport matching against offline demonstrations as the reward. BeT~\cite{shafiullah2022behavior} introduces a residual action corrector that refines continuous actions on top of a discretized imitation policy. IBRL~\cite{hu2023imitation} does not directly use Residual RL, but it bootstraps an RL policy from an IL policy by using IL actions as alternative proposals for both online exploration and critic updates. Closest to our method is Policy Decorator~\cite{yuan2024policy}, which learns bounded residual actions using controlled exploration. Unlike Policy Decorator, which uniformly samples actions from the base and residual policies, we use uncertainty estimates of the base policy to decide when to learn and apply corrective residual actions. 

\textbf{Uncertainty Estimates in Imitation Learning.} 
Uncertainty estimation plays a crucial role in improving the reliability and robustness of machine learning models, particularly in decision-making and RL. Various approaches have been proposed to quantify uncertainty, including distance-based techniques \cite{permenter2023interpreting,suh2023fighting}, ensemble-based techniques~\cite{an2021uncertainty, sekar2020planning,georgakis2022uncertainty}, and learning another model to estimate uncertainty~\cite{chan2024estimating,bucher2021adversarial,gummadi2024metacognitive}. SGP \cite{suh2023fighting} proposes a method that measures the distance of a given input to the training data distribution. This approach assumes that samples farther from the training distribution exhibit higher uncertainty, making it particularly useful for detecting out-of-distribution (OOD) inputs and improving model generalization. Diff-DAgger \cite{lee2024diff} uses the loss function of a Diffusion model to estimate uncertainty, where a higher loss indicates greater uncertainty. Our algorithm is agnostic to the uncertainty quantification method, and we test our approach with different uncertainty metrics.

\section{Uncertainty aware Residual RL for stochastic policies}
We describe how to incorporate uncertainty estimates in Residual RL in Sec.~\ref{sec:uncertainty} and our modified off-policy RL algorithm in Sec.~\ref{sec:combined_action}. 

\subsection{Uncertainty Aware Residual RL}
\label{sec:uncertainty}
Prior works in Residual RL suffer from unrestrained exploration as they learn corrective residual actions uniformly over the entire state space. Our key insight is to improve exploration by focusing residual learning on regions in which the base policy is not confident. We propose using the uncertainty of the base policy to decide when to learn a residual action for the base policy. If the base policy is certain about its action for the current state, we directly use the base policy action$a_b$ in the environment; we only use a corrective residual action $a_r$ when the base policy is uncertain. 
Our proposed approach is agnostic to the uncertainty quantification method, and we demonstrate our method with two distinct uncertainty metrics: distance-to-data and ensemble variance.
Distance-to-data has been used to calibrate the uncertainty of a model by measuring how out-of-distribution the current state is from an existing dataset \cite{suh2023fighting}. For a dataset $D$ where each state has $F$ features, we estimate uncertainty using the minimum L2 norm of the current state $s$ to all states $d \in D$ :
\begin{equation}\label{eqn:uncertainty_estimation}
\text{uncertainty}_{d}(s) = \min_{d\in D} \sqrt{\sum_{i=1}^{F} (d_i - s_i})^2 \ .
\end{equation}
Another popular approach to estimating uncertainty is measuring the variance in predicted actions among an ensemble of policies. For an ensemble of $N$ base policies $\pi_b \in \pi_{B}$, the ensemble uncertainty can be defined as:
\begin{equation}
    \text{uncertainty}_{e}(s) = \frac{1}{N} \sum_{b=1}^N \left( \pi_b(a \mid s) - \frac{1}{N} \sum_{i=1}^N \pi_i(a \mid s) \right)^2 \ .
\end{equation}
We can use our desired uncertainty metric with a threshold $\tau$ to measure the confidence of the base policy. This can be formulated as 
\begin{equation} \label{eqn:uncertainty_threshold}
    a_{\text{taken}} =
    \begin{cases}
       a_{\text{b}} & \text{if uncertainty} < \tau \\
    
       a_{\text{b}} + a_{\text{r}} & \text{otherwise}\ .
    \end{cases}
\end{equation}

As learning progresses, we decay this uncertainty threshold $\tau$ exponentially from a maximum uncertainty threshold value $U$ according to the following equation: 

\begin{equation} \label{eqn:threshold_decay}
    \tau = U * e^{\frac{-\text{step}}{\text{decay\ rate}}}\ .
\end{equation}

The uncertainty threshold $\tau$ ultimately decays to 0 to let the residual policy take over. We perform ablations for different decaying strategies in Sec.~\ref{sec:ablation_decay_strategy}.


\subsection{Optimizing Residual RL for Stochastic Policies}
\label{sec:combined_action}
The original Residual RL algorithms are formulated to learn only using partial residual actions, operating under the assumption that the underlying base policy is deterministic and can be implicitly inferred. Therefore, it learns the $Q$ function for only the partial residual action, which is different from the action taken in the environment:
\begin{equation}
    Q(s,a_{r}) = \mathbb{E}_{s' \sim P} \left[ R(s, a_r, s') + \gamma V^\pi(s') \right].
\end{equation}
Incorporating stochastic policies into the residual transition function $T_r$ can make it much harder to learn, as they are noisier in their predictions and hence difficult to model. Thus, we suggest using Eq.~\ref{eqn:transition} to retreat back to the original MDP formulation. Consequently, the $Q$ function is learned for the combined action $a_c$, which is the actual action used during environment interaction :
\begin{equation}
    Q(s,a_{c}) = \mathbb{E}_{s' \sim P} \left[ R(s, a_b + a_r, s') + \gamma V^\pi(s') \right]
\end{equation}
Previously, ResiP \cite{ankile2024imitation} proposed augmenting the observed state with base action to provide the missing information for on-policy RL. We instead propose learning the critic for the combined action $a_c$ for off-policy RL, providing the necessary information about the base policy to the $Q$ function while also making it invariant to the split between the residual and base actions. Specifically, we modify Soft Actor-Critic~\cite{haarnoja2018soft} in the following ways (changes marked in \teal{green}). Initially, we store both the base action $a_b$ and the combined action $a_c$ in the replay buffer. While computing the target values, we add the base action $a_b$ to the residual action $a_r$ sampled from the actor:  
\begin{equation}\label{eqn:update_target}
\begin{split}
    y(r,s',d) &= r + \gamma (1 - d)\  * \\
     & \Big[ \min_{i=1,2} Q_{\phi'_i}(s', \teal{a'_{r} + {a'_b}}) - \alpha \log \pi_{r}(\teal{a'_{r}} | s') \Big], \\
     &\quad\teal{a'_{r} \sim \pi_{r}(\cdot | s')}.
    \end{split}
\end{equation}
When updating the $Q$ function, we use the combined action stored in the replay buffer:
\begin{equation}\label{eqn:update_Q}
        J_Q(\phi_i) = \mathbb{E} \left[ \left(Q_{\phi_i}(s, \teal{a_c}) - y(r,s',d) \right)^2 \right], \quad i=1,2 \ .
\end{equation}
When updating the actor, we again add the base action to get the $Q$ value:

\begin{equation}
\begin{split}\label{eqn:sample_actor}
        J_{\pi}(\theta) &= \mathbb{E} \left[ Q_{\phi_i}(s, \teal{a_{r} + a_{b}})  - \alpha \log \pi_{r}(\teal{a_{r}} | s)\right] , \\
        &\quad i=1,2 , \teal{\quad a_{r} \sim \pi_{r}(\cdot | s)} \ .
\end{split}
\end{equation}
The complete algorithm is described in Alg.~\ref{alg:one} with the proposed changes from SAC marked in \teal{green}.

\begin{minipage}{0.95\linewidth}
\begin{algorithm}[H]
\small
\captionsetup{font=small}
{\caption{Uncertainty aware Residual RL}\label{alg:one}
\begin{algorithmic}[1]
\State Initialize parameters of \teal{residual policy $\pi_r$}, Q-functions $Q_{\phi_1}, Q_{\phi_2}$, and temperature $\alpha$
\State Initialize target Q-function parameters $\phi'_1 \leftarrow \phi_1$, $\phi'_2 \leftarrow \phi_2$
\teal{\State Initialize base policy $\pi_b$}
\For{each environment step}
    \State Sample residual action from actor \\ \hspace{1cm} $a_r \sim \pi_r(s_t)$
    \teal{\State Sample base action from base policy \\ \hspace{1cm} $a_b \sim \pi_b(s_t)$}

    \teal{\State Calculate uncertainty threshold $\tau$ for the base policy,  Eq.~\ref{eqn:threshold_decay}}
    \teal{\State Calculate the uncertainty in base policy, Eq.~\ref{eqn:uncertainty_estimation}}
    \teal{\State Select the action to be taken in the environment,  Eq.~\ref{eqn:uncertainty_threshold}}
    \State Observe next state $s_{t+1}$ and reward $r_t$
    \teal{\State Store $(s_t, a_c, a_b, r_t, s_{t+1})$ in replay buffer $\mathcal{D}$}
\EndFor
\For{each gradient update step}
    \State Sample a minibatch $\{(s, a_c, a_b, r, s')\}$ from $\mathcal{D}$
    \State Compute target value \teal{according to Eq~\ref{eqn:update_target}}
    \State Update Q-functions by minimizing \teal{according to Eq.~\ref{eqn:update_Q}}
    \State Update policy $\pi_{\theta}$ \teal{using Eq.~\ref{eqn:sample_actor}}
    \State Update target networks $\phi'_i$
\EndFor
\end{algorithmic}
}
\end{algorithm}
\end{minipage}

\section{Experiments}
In this section we describe our experiment setup, baselines, results and ablations. 

\subsection{Experiment Setup} \label{sec:exp_setup}

\subsubsection{Environments}\label{sec:environments}

\begin{figure}
    \centering
    \includegraphics[width=0.9\linewidth]{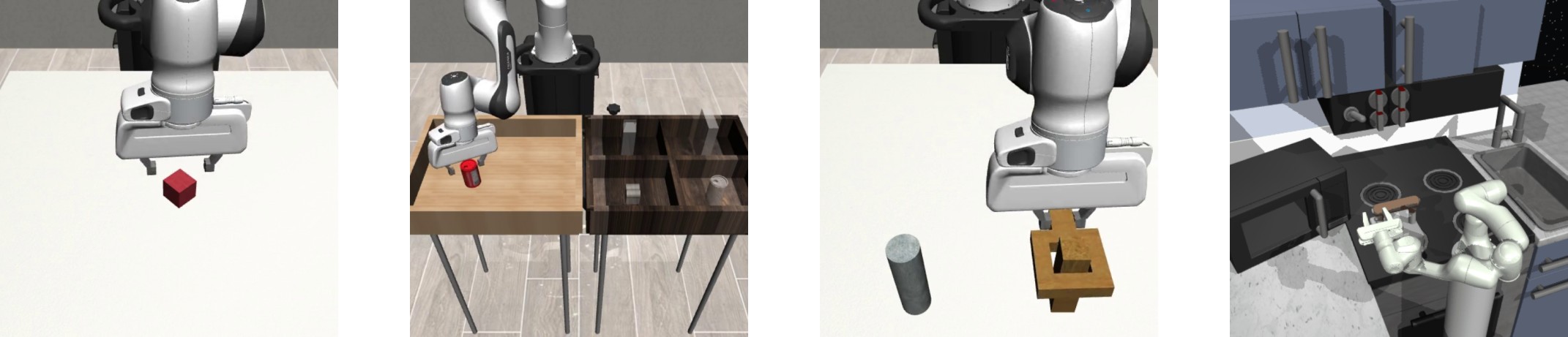}
    \caption{We test our proposed approach on the \emph{Lift}, \emph{Can}, and \Square tasks from Robosuite \cite{mandlekar2021matters} and the \emph{Franka Kitchen} Task from D4RL \cite{fu2020d4rl}.}
    \label{fig:task_visualization}
    \vspace{-1.5em}
\end{figure}

We conduct evaluations on the environments visualized in Fig.~\ref{fig:task_visualization} and described below.
All environments use sparse rewards and state-based observations. 

\textbf{Robosuite}: We evaluate on three tasks from the Robosuite simulator~\cite{mandlekar2021matters}: 1) \textbf{Lift}, where the robot arm must pick up a block placed at a random initial position on the table. 2) \textbf{Can},  where the robot arm has to pick up a can from one table and place it in the top right corner of another table. 3) \textbf{Square}, where the robot arm must pick up a square nut from the table and place it onto a square bolt.

\textbf{Franka Kitchen}: The \emph{Franka Kitchen} environment from the D4RL benchmark \cite{fu2020d4rl} features a Franka robot that is required to interact with various objects to achieve a multitask goal configuration. It receives a reward of 1 for successfully completing each of the 4 sub-goals, and we report the normalized reward for each trajectory. The environment includes three datasets for the \emph{Franka Kitchen} task from the D4RL benchmark: 1) \textbf{Kitchen Complete}: This dataset is limited in size and consists solely of positive demonstrations. We perform additional experiments with the other two datasets. 2) \textbf{Kitchen Mixed}: This dataset includes undirected demonstrations, with a portion of them successfully solving the task. 3) \textbf{Kitchen Partial}: This dataset also contains undirected demonstrations, but none of them fully solve the task. However, each demonstration successfully addresses certain components of the task.
\subsubsection{Base Policies}
We consider two kinds of IL base policies to test the robustness of our algorithm.
\begin{figure*}
\centering
\includegraphics[width=0.9\linewidth]{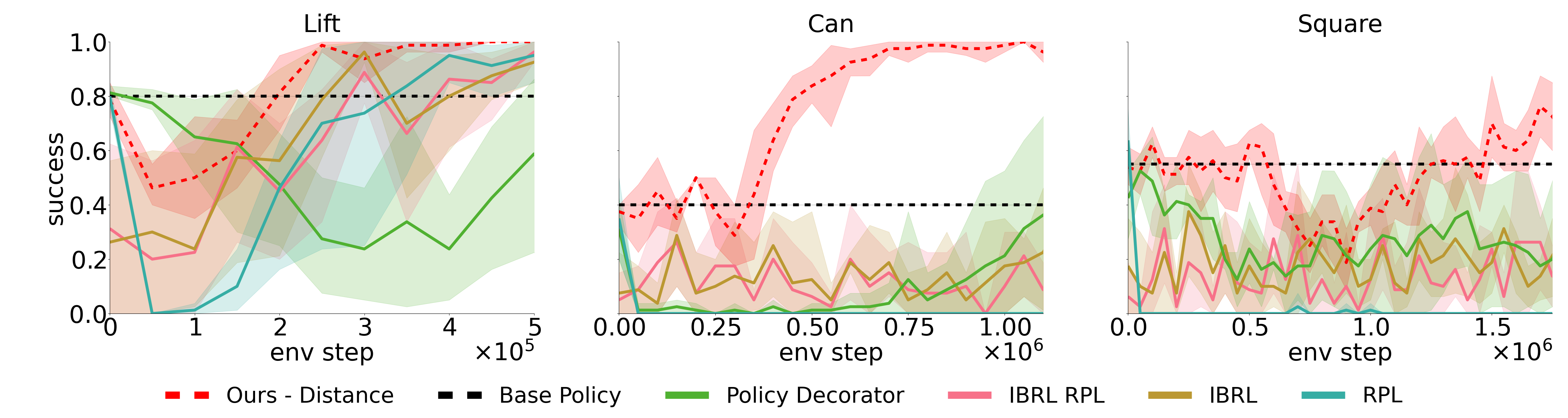}
\caption{Results on Robosuite environments with a GMM base policy. Our method is able to outperform all other baselines in all tasks. The error bars indicate 95\% confidence interval.}
\label{fig:robomimic}
\end{figure*}
\begin{figure*}
\centering
\includegraphics[width=0.9\linewidth]{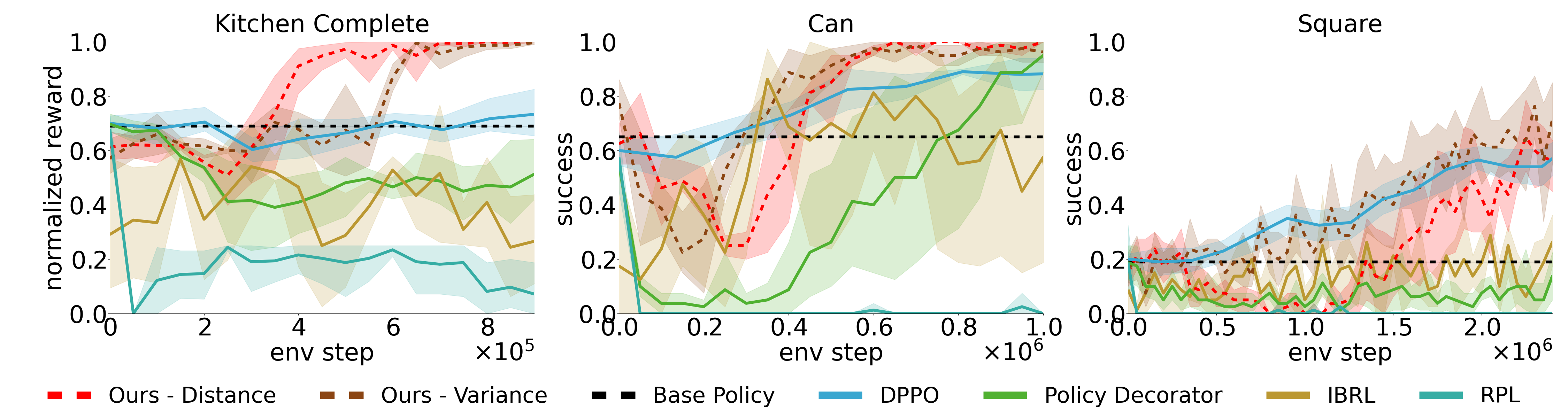}
\caption{Results on Franka Kitchen and Robosuite environments with a Diffusion base policy. Our method is able to outperform all baselines for \emph{Kitchen Complete} and \emph{Can} task, and has comparable performance for \emph{Square} Task. The error bars indicate 95\% confidence interval.}
\label{fig:diffusion}
\vspace{-1.5em}
\end{figure*}

\textbf{Gaussian Mixture Model-based policy}: We utilize GMM-based behavior cloning (BC) policies from \cite{mandlekar2021matters}, which has an RNN backbone. To introduce an additional challenge, we train policies for the \Lift and \Can tasks using noisier multi-human demonstrations. Since the \Square task is inherently challenging, we use proficient-human demonstrations, as even high-quality demonstrations struggle to completely solve the task. 

\textbf{Diffusion-based policy}: We also test our method with a diffusion base policy \cite{chi2024diffusionpolicy}. To ensure a consistent comparison, we use the same Diffusion policies from DPPO \cite{ren2024diffusion} for our experiments. For Robosuite tasks, they use noisier multi-human demonstrations, while for the \emph{Franka Kitchen} environments, they directly use datasets from D4RL. 

\subsection{Baselines} \label{sec:baselines}
\textbf{Finetuning methods}: As a baseline for Diffusion-based policies, we use the recently proposed \textbf{DPPO} \cite{ren2024diffusion}. DPPO formulates the denoising process of the Diffusion policy as a separate MDP, effectively modeling the entire trajectory as a sequence of MDPs. The policy is then optimized using policy gradients across this entire chain of MDPs. \\
\textbf{Demo augmented RL methods}: We compare our method against two demo-augmented RL approaches. \textbf{IBRL} \cite{hu2023imitation} maintains both an IL policy trained on demonstrations and an RL policy trained from scratch. During environment interaction and RL training, both policies propose actions, and the action with the highest $Q$ value is selected. A variant of this approach, \textbf{IBRL-RPL}, replaces the RL policy learned from scratch with a residual policy. We conduct experiments with both versions of IBRL.\\
\textbf{Residual RL methods}: We test our method against the most closely related approach and current state-of-the-art residual learning algorithm, \textbf{Policy Decorator} \cite{yuan2024policy}, which also aims to mitigate excessive exploration in Residual RL. It addresses this by sampling actions uniformly from both the residual and base policies with a progressive exploration schedule. Additionally, it bounds the actions of the residual policy. For completeness, we also compare our approach with the standard \textbf{Residual RL} \cite{silver2018residual, johannink2019residual} algorithm, incorporating our proposed modifications to the critic without using uncertainty estimation.

\subsection{Results and Analysis}
\label{sec:results}

We evaluate all experiments with 5 seeds over 20 runs and provide hyperparameters in Appendix~\ref{sec:hyperparameters}.

\subsubsection{GMM-based policies}\label{sec:gmm_results}
We present the results of our experiments with GMM-based policies in Figure~\ref{fig:robomimic}. We plot the success rate over the course of environment interactions. Our method with distance to data metric (\red{red} line) outperforms all the baselines in the three Robosuite environments. We note that there is an initial dip in the performance for our method where the residual policy is in the exploration phase, but it starts improving once the exploration phase ends. IBRL performs the best out of the other baselines, reaching comparable performance to our method for the \emph{Lift} Task, though the initial dip in performance is more significant, and it is still unstable afterwards. We performed a hyperparameter sweep for the two additional parameters of Policy Decorator, namely the residual bound and the decay rate with more details in the Appendix D. Our method converges in the same number of timesteps, while Policy Decorator's performance is still improving. We suspect it is due to the more targeted exploration of our method using uncertainty estimates. The standard Residual Policy Learning method only improves over the base policy performance for the Lift Task.

\subsubsection{Diffusion policies} \label{sec:diffusion_results}
We present the results for Diffusion policies in Figure~\ref{fig:diffusion}. We run the experiments for the \emph{Kitchen Complete} environment of the D4RL benchmark. In Robosuite, we perform experiments in the \Can and \Square environments, excluding the \Lift environment because of the near-optimal performance of Diffusion policy on that task. We performed a similar hyperparameter sweep of Policy Decorator as mentioned in Sec.~\ref{sec:gmm_results} with details in Appendix D. Similarly, our method is able to outperform Policy Decorator in all environments. Our approach is able to achieve higher success rates than all the baselines in the \emph{Kitchen Complete} environment with both types of uncertainty metrics. We note that even though DPPO's performance is stable in \emph{Kitchen Complete} and \Can environments, its improvement over the initial performance of the base policy is slow compared to our approach, despite an initial dip in the performance. Our method with the ensemble variance metric has comparable performance to DPPO for the \Square task. These results suggest that our approach is most promising when the initial base policy performance is average, and is comparable in scenarios where the initial base policy performance is bad. Also, we note that distance-to-data as a metric works better than ensemble variance for the  \emph{Kitchen Complete} environment but not the \Can and \Square environment. We hypothesize that this is because of the high quality demonstrations in the \emph{Kitchen Complete} environment, whereas the noisier multi-human demonstrations of \Can and \Square tasks result in a noisier distance-to-data metric. Nevertheless, our method converges for both the metrics.


\subsection{Combined Action vs. Residual Action} \label{sec:ablation_complete_action}

\begin{figure}[h]
\vspace{0.5em}
    \centering
    \includegraphics[width=0.9\linewidth]{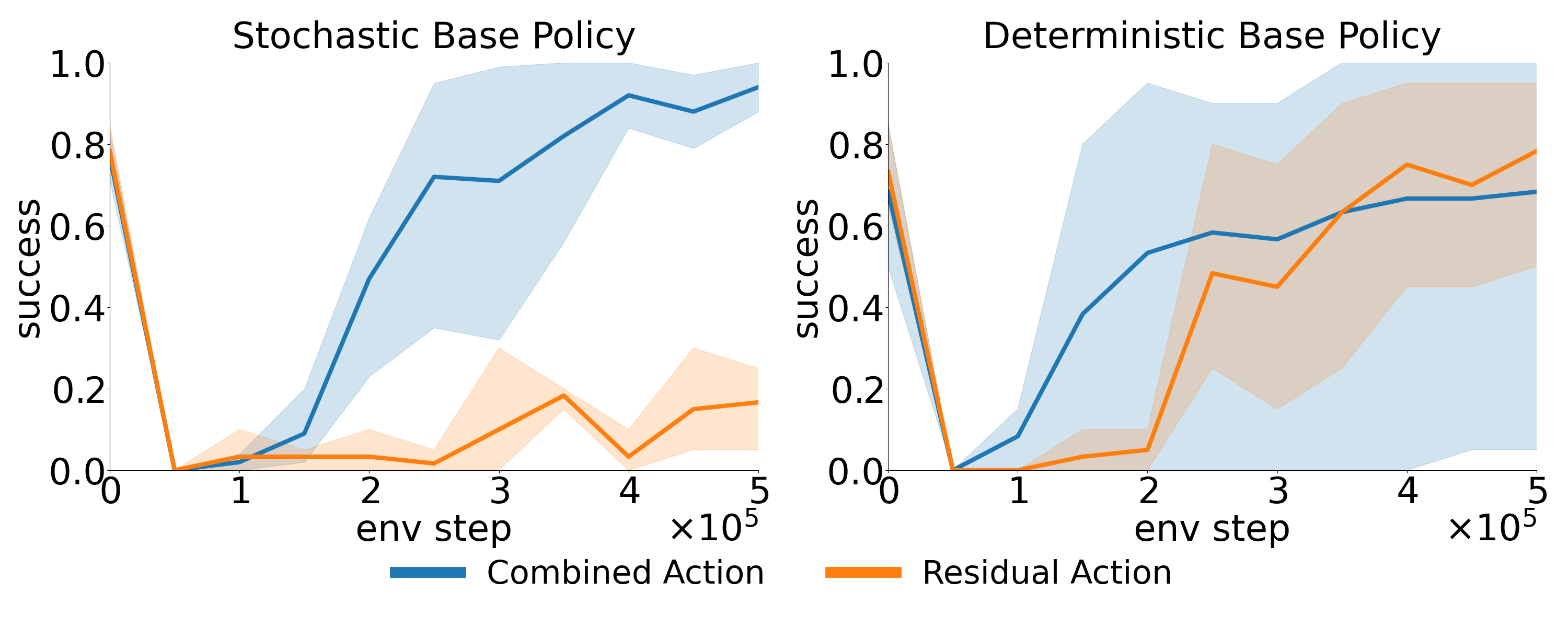}
    \caption{Learning with either the residual or the complete action works well with deterministic base policies, but learning with complete action is required for stochastic base policies.}
    \label{fig:complete_partial}
\end{figure}

To emphasize the significance of utilizing combined actions for stochastic policies, we compared our modified SAC algorithm for Residual RL \textit{without} uncertainty estimates against the original Residual RL formulation. For this comparison, we used the GMM policy as the stochastic base policy and a standard MLP policy as the deterministic base policy, applied to the \Lift task in Robosuite. The results for both stochastic and deterministic base policies are shown in Figure~\ref{fig:complete_partial}. The findings reveal that relying solely on the residual action does not yield effective results for stochastic base policies, highlighting the necessity of using our combined action formulation in such cases. In contrast, for deterministic base policies, either residual actions or combined actions can be used.

\begin{figure}[h]
\centering
    \includegraphics[width=0.45\linewidth]{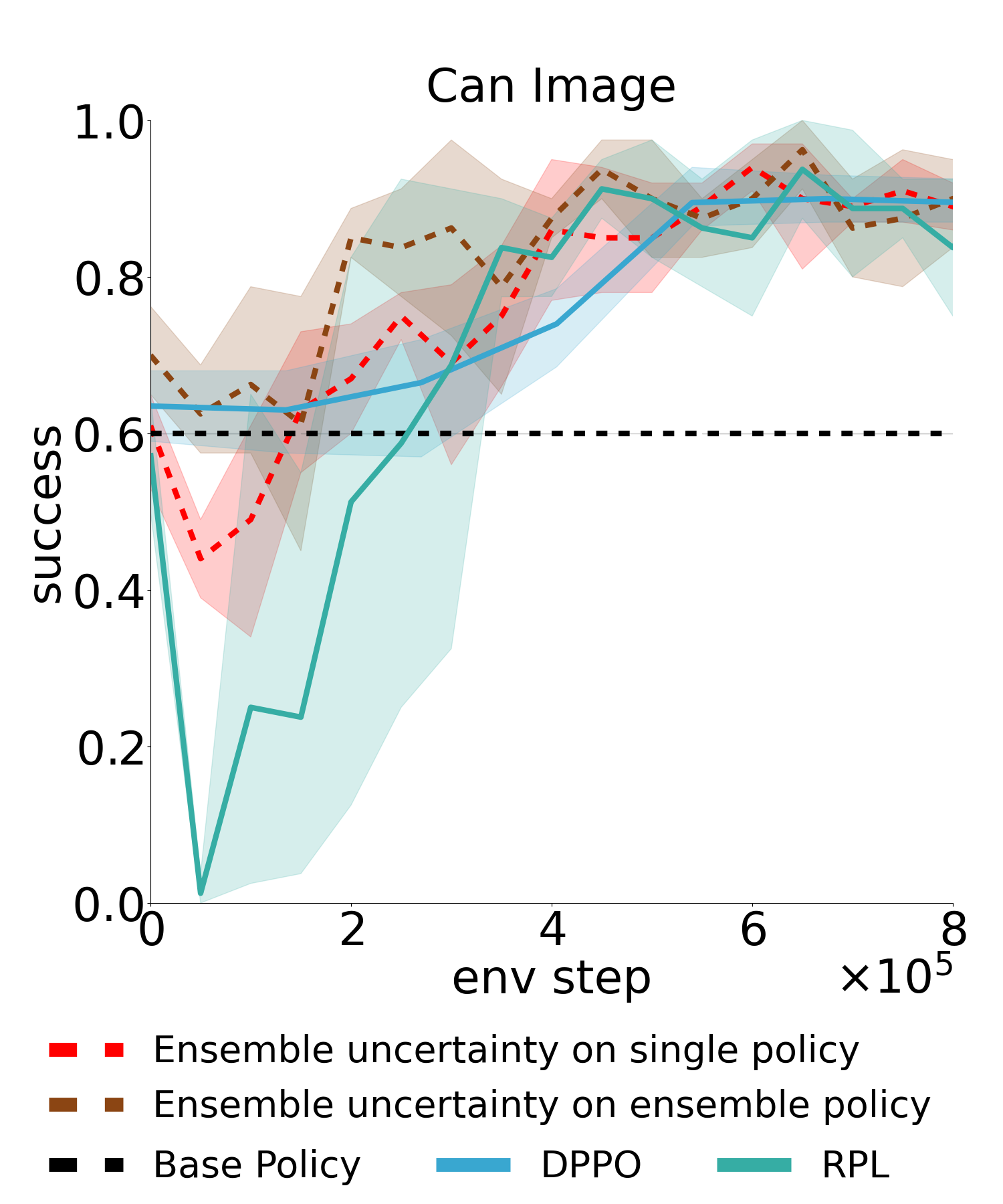}
    \caption{Results on the Can-Image Task.}
    \label{fig:image_based}
\end{figure}
\subsection{Image Based Experiments} \label{sec:image_experiments}
We evaluated our algorithm on the \Can task with image observations. Given that distance-to-data is a less reliable metric in high dimensional input spaces such as images, we instead used ensemble variance to quantify uncertainty. We explored two strategies: first, using ensemble variance to enhance the performance of a single policy, and second, leveraging it to improve an ensemble of policies. Our method was compared against DPPO, the strongest baseline for finetuning diffusion policies, and Residual Policy Learning (RPL) without uncertainty estimation. The results of our algorithm on the image-based \Can Task are shown in Fig~\ref{fig:image_based}. The results indicate that the ensemble-based approach starts with strong performance and avoids early dips, likely due to the robustness offered by ensembling. In contrast, the naive RPL strategy initially crashes to 0\% performance after the initial online learning steps, though it recovers over time. DPPO, as observed previously, maintains steady but slow improvement. These findings show our approach also performs well in image-based environments.

\begin{figure}[t]
    \centering
    \includegraphics[width=0.7\linewidth]{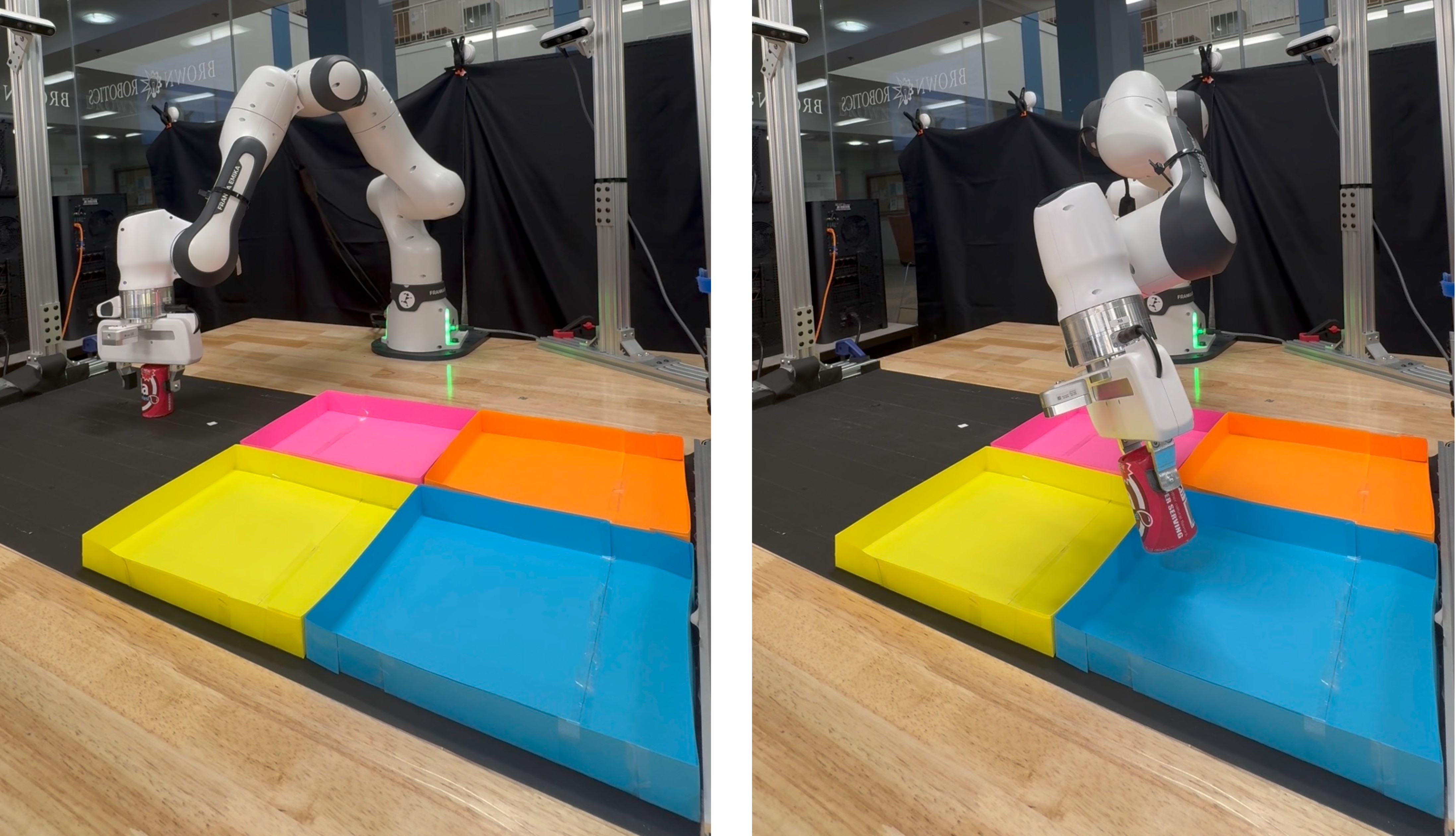}
    \caption{Sim-to-Real setup for the Robomimic Can task.}
    \label{fig:real_world}
\end{figure}

\begin{figure}[t]
\centering
\includegraphics[width=\linewidth]{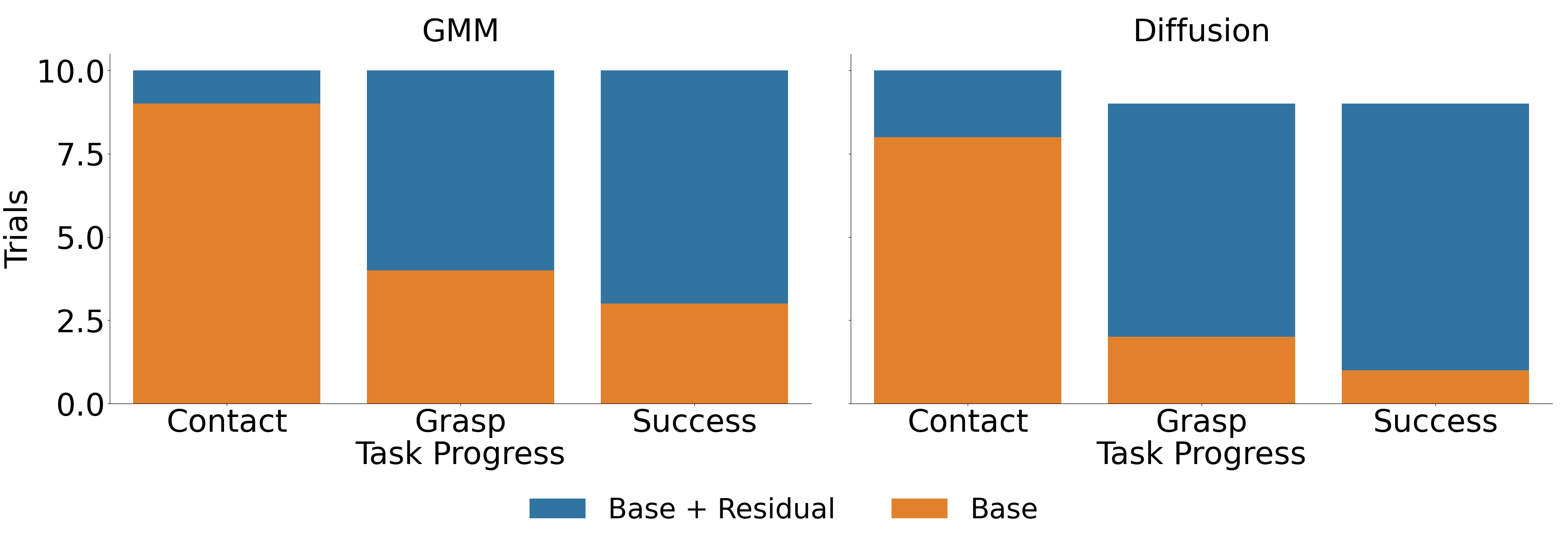}
\caption{Results for the policy deployment in real world.}
\label{fig:result_real_world}
\vspace{-1em}
\end{figure}
\subsection{Real World Experiments} \label{sec:real_world_experiments}
We deployed our learned policy for the \Can task in the real world in a zero-shot setup as shown in Fig~\ref{fig:real_world}. The task is set up similarly to the simulated version, where the robot has to pick up the can from one side of the table and place it in the blue bin on the other side of the table. We used Grounded SAM \cite{ren2024grounded} and Foundation Pose \cite{wen2024foundationpose} to get the real-time state of the can object. We evaluated four policies in our real-world setup: 1) GMM base, 2) GMM base + Residual, 3) Diffusion base, and 4) Diffusion base + Residual. We compare the performance in terms of task progression divided into three stages: contacting, grasping, and placing. Fig~\ref{fig:result_real_world} shows the performance of each policy across 10 trials. Policies learned with Residual RL retain nearly all of their original performance in simulation without any domain randomization, whereas base policies struggled. This supports the observation that policies trained using RL tend to be more robust than behavior cloning policies, as they benefit from richer interaction with the environment.

\subsection{Ablations} \label{sec:ablations}

\begin{figure}
\centering
\includegraphics[width=\linewidth]{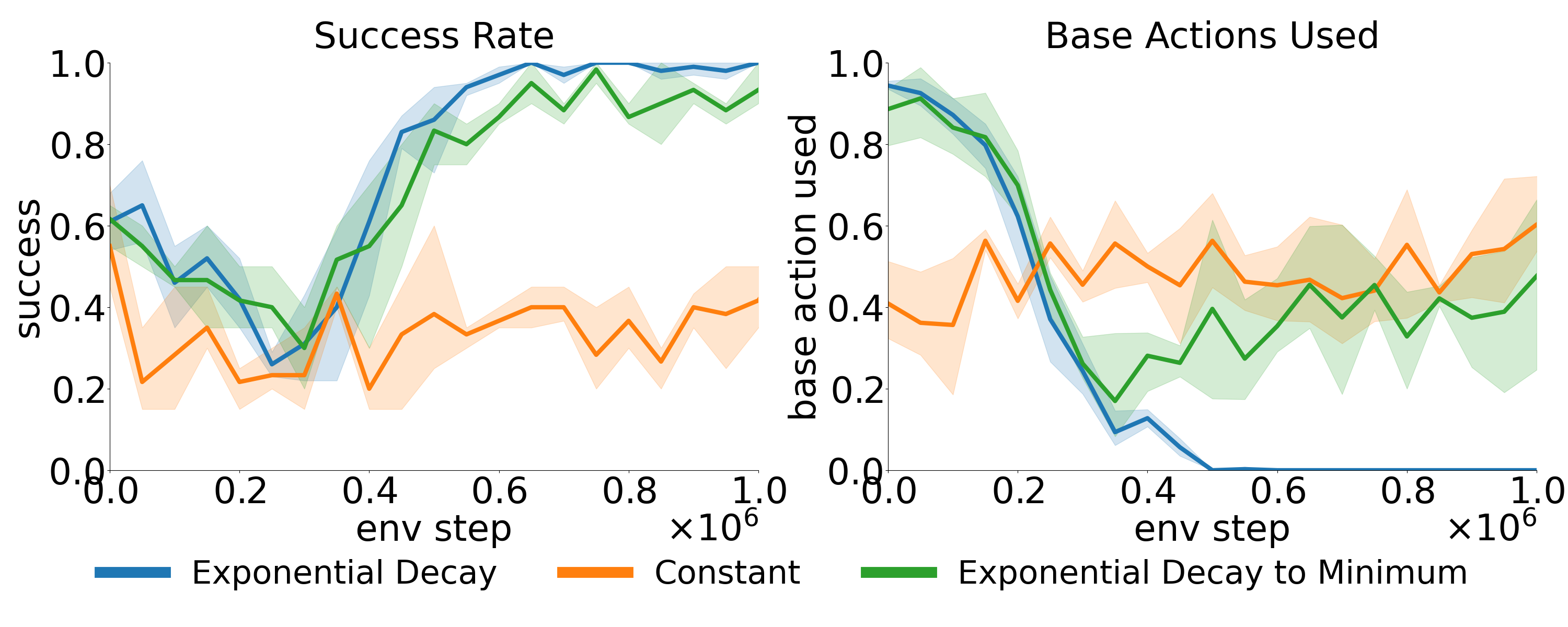}
\caption{Ablations for different threshold decay strategies. Decaying exponentially performs best.}
\label{fig:type_threshold}
\end{figure}

\subsubsection{Threshold Decay Strategy} \label{sec:ablation_decay_strategy}
We tried different strategies for decaying the uncertainty threshold $\tau$: exponentially decaying the threshold to zero, exponentially decaying to a minimum threshold, and keeping the threshold constant. We plot success rate and the percentage of base policy actions used in Figure~\ref{fig:type_threshold}. We observe that exponential decay has the most stable performance out of the three. Exponentially decaying to a minimum threshold converges at a lower success rate compared to exponentially decaying the threshold to 0. The base actions used when decaying the threshold to a minimum value start increasing once that minimum threshold value is reached, signifying that the optimal policy stays within the distribution of the base policy. Keeping the threshold value constant restricts the residual policy from escaping the performance of the initial base policy. 

\begin{figure}
\begin{minipage}{0.45\textwidth}
    \centering
    \includegraphics[width=0.45\textwidth]{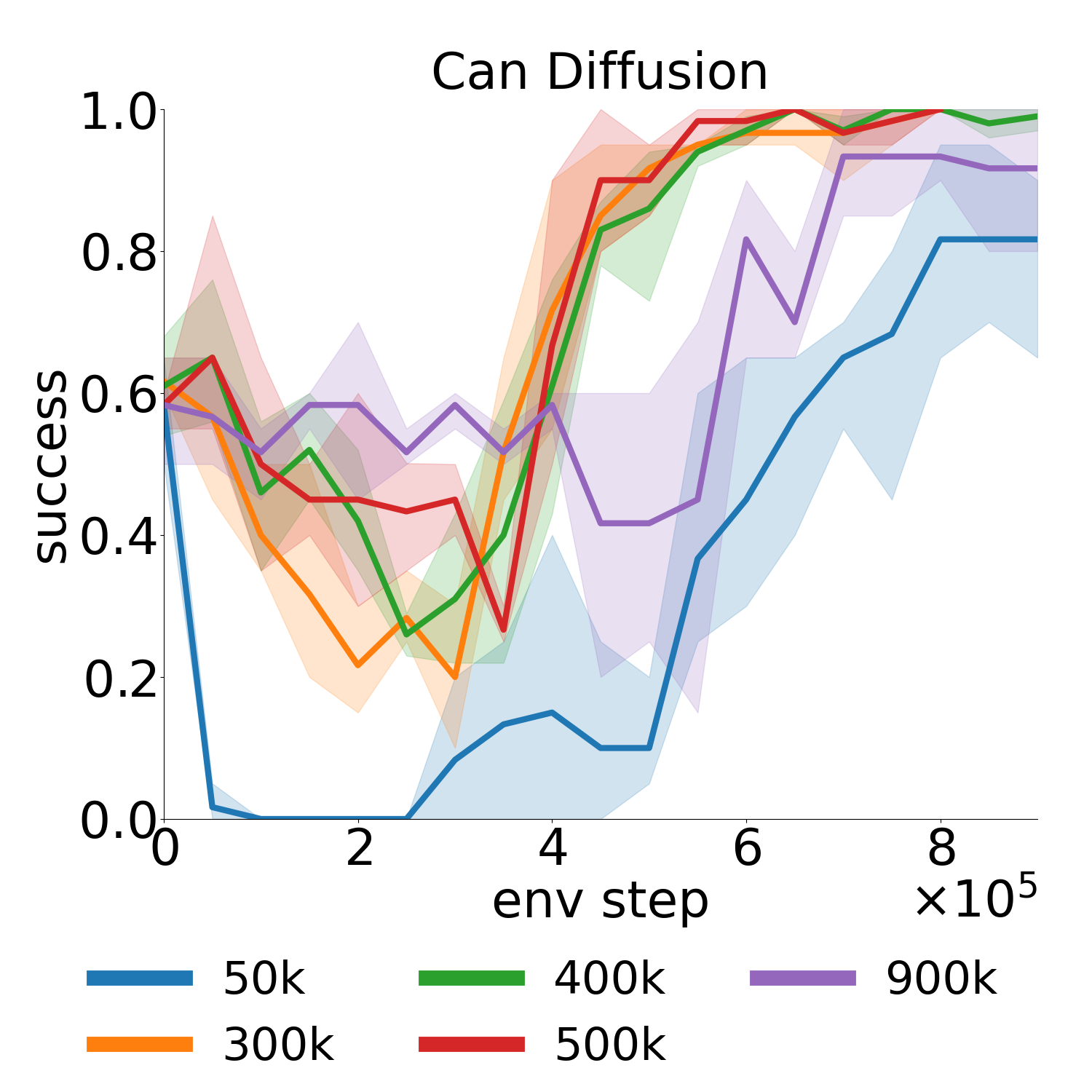}
    \caption{Higher decay rates result in slower convergence whereas lower decay rates results in aggressive exploration.}
    \label{fig:decay_rates}
\end{minipage}
\begin{minipage}{0.45\textwidth}
\centering
\includegraphics[width=\textwidth]{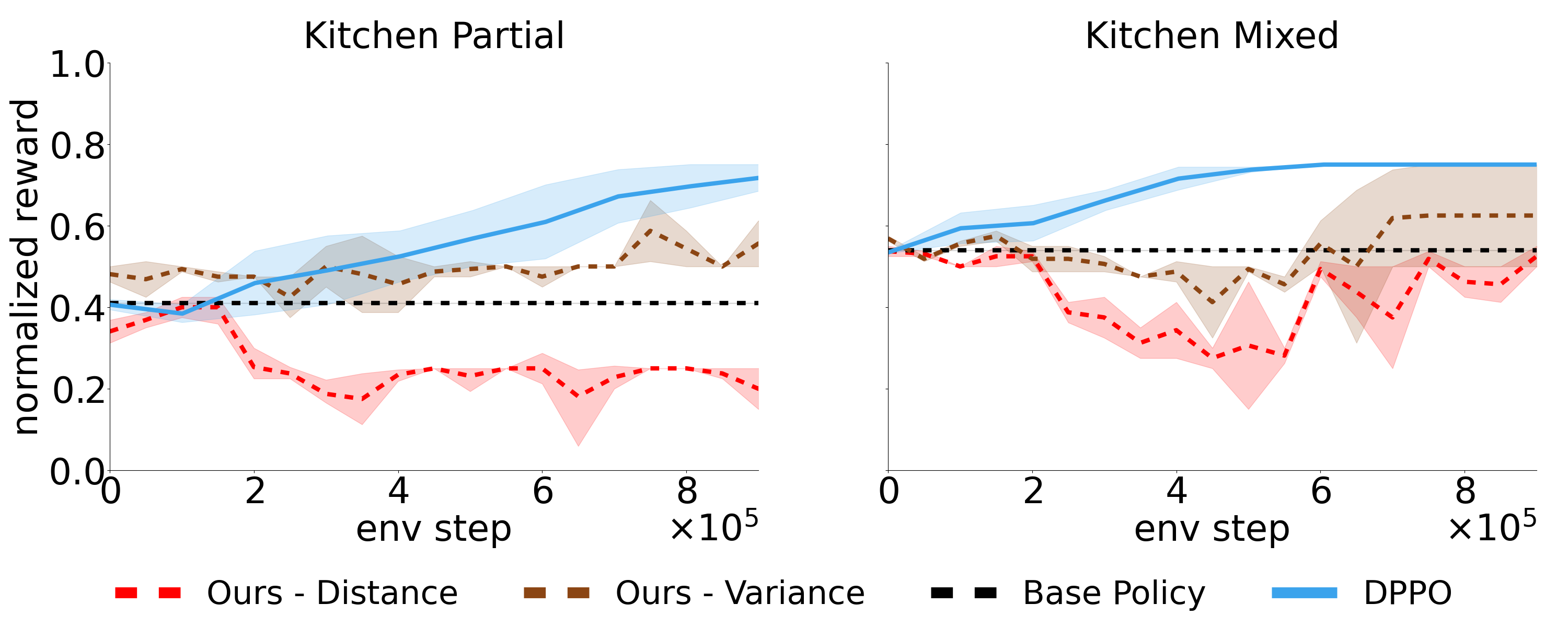}
\caption{Ensemble variance is a better uncertainty metric for Kitchen Partial and Kitchen Mixed environments as the training data also contains random trajectories.}
\label{fig:kitchen_ablation}
\end{minipage}
\vspace{-1em}
\end{figure}



\subsubsection{Threshold Decay Rate} \label{sec:ablation_decay_rates}
We evaluated our algorithm with different decay rates in Figure~\ref{fig:decay_rates}. Lower rates resulted in aggressive exploration, causing performance dips that were hard to recover from, while higher rates slowed the convergence. We need to balance decay rates for effective exploration without hindering convergence. However, our approach is robust to the chosen decay rate, as the performance is similar for decay rates ranging between 300k and 500k.

\subsubsection{Kitchen Environments} \label{sec:ablation_kitchen}
We tested our method on the other two variations of the kitchen environments, \emph{Kitchen Partial} and \emph{Kitchen Mixed}. As seen in Figure~\ref{fig:kitchen_ablation}, our method with the distance to data metric is not able to outperform the base policy. One key assumption we make in our algorithm is that when the base policy is certain, it will also be correct. This assumption does not hold true for these two environments: the base policies for these two environments are trained on the random play data, so the confidence and correctness of the policy are not strongly correlated, resulting in poorer performance. However, ensemble variance is able to provide a better uncertainty metric when random play data is involved and results in improved performance.

\section{Conclusion and Future Work}
We propose two improvements to the Residual RL framework to accelerate the learning with stochastic base policies. First, we use uncertainty estimates of the base policy to constrain exploration in residual learning. Second, we adapt Residual RL to handle stochastic base policies by proposing an asymmetric actor-critic approach in which the critic observes the combined action, while the actor predicts only the residual action. While our proposed method demonstrates strong performance, it would also benefit from a more robust epistemic uncertainty metric. We believe that, with reliable uncertainty metrics, our approach could also be applied to larger models including robot foundation models. In the future, we would also like to dynamically set the values of uncertainty threshold decay based on the performance of the base policy.

\section*{Acknowledgments}
LD was supported by the NASA-Kennedy award ``Efficient and High Quality Space Robotic VR Teleoperation through Neural Scene and Object Reconstruction." under award number 80NSSC23M0075. SV and GK was supported by the Office of Naval Research (ONR) under REPRISM MURI N000142412603 and ONR grant \#N00014-22-1-2592, as well as by the National Science Foundation (NSF) via grant \#1955361. Partial funding for SV and GK was also provided by the Robotics and AI Institute. We would also like to thank Ondrej Biza and Ivy He for helpful discussions.


{\appendices
\section*{A. Environments}
The observation space of each environment used is below:
\begin{enumerate}
    \item \textbf{Lift} - 19 dim obs space with object state and robot eef.
    \item \textbf{Can} - 23 dim obs space with object state and robot eef.
    \item \textbf{Square} - 23 dim obs space with object state and robot eef.
    \item \textbf{Kitchen} - 60 dim obs space with all object states and velocities, and robot joint states and angular velocity.
\end{enumerate}
Action space for all Robosuite environments is 7 DoF end effector pose while for Kitchen environment is the 9 DoF joint angular velocity and gripper linear velocity.
\section*{B. Base Policies}
\textbf{GMM Base Policy} - We use Robomimic \cite{mandlekar2021matters} to train Gaussian Mixture Model based policies with a Recurrent Neural Network backbone.

\textbf{Diffusion policy} - We used the same base policies from DPPO \cite{ren2024diffusion} to keep the comparisons consistent. The Diffusion policies are trained with an action horizon of 1 and 20 denoising steps.

\section*{C. Hyperparameters} \label{sec:hyperparameters}
We used the same hyperparameters in each environment for both GMM-based and Diffusion-based. We keep the same parameters for actor and critic for Robosuite environments and use the advised hyperparameters from the DPPO paper for the kitchen environment. Hyperparameter details can be found in Table~\ref{tab:hyperparameters}. The uncertainty threshold value $U$ and decay rate values for our proposed approach with distance-to-data and ensemble variance metric can be found in Table~\ref{tab:alpha_decay_rate}.
\begin{table}[h]
\vspace{1.5em}
    \centering
    \begin{tabular}{|c|c|c|c|c|}
         \hline
         \textbf{Environment} & \textbf{Actor \& Critic Dimensions}& \textbf{Actor lr} & \textbf{Critic lr} \\
         \hline
         Robosuite & (256,256) & 1e-4 & 1e-4 \\
         \hline
         Kitchen & (256,256,256) & 1e-5 & 1e-3 \\
         \hline
    \end{tabular}
    \caption{Hyperparameters used for each environment}
    \label{tab:hyperparameters}
\end{table}

\begin{table}[h]
    \centering
    \begin{tabular}{|c|c|c|c|}
         \hline
         \textbf{Environment} &\textbf{Base Policy} & $\textbf{U}$ & \textbf{Decay Rate} \\
         \hline
         \multicolumn{4}{|c|}{Distance-to-data} \\
         \hline
         Lift & GMM & 1e-6  & 200k  \\
         \hline
         Can & GMM & 2e-5  & 75k \\
         \hline
         Square & GMM & 5e-5 & 150k  \\
         \hline
         Kitchen Complete & Diffusion & 2.5e-3 & 200k  \\
         \hline
         Can & Diffusion & 4.5e-5 & 400k \\
         \hline
         Square & Diffusion & 4.5e-5 & 1M  \\
         \hline
         \multicolumn{4}{|c|}{Ensemble variance} \\
         \hline
         Kitchen Complete & Diffusion & 0.5 & 200k  \\
         \hline
         Can & Diffusion & 0.2 & 500k \\
         \hline
         Square & Diffusion & 0.4 & 750k  \\
         \hline
         
    \end{tabular}
    \caption{Uncertainty threshold $U$ and Decay Rate values for distance-to-data and ensemble variance.}
    \label{tab:alpha_decay_rate}
\vspace{-1em}
\end{table}
         
    

\section*{D. Tuning Policy Decorator}
Policy Decorator ~\cite{yuan2024policy} has two hyperparameters namely, $\alpha$ the residual bound which scales the residual action to limit exploration and $H$ to schedule the exploration progressively. According to their paper, the $\alpha$ value is set close to the action scale of demonstration data while it is advised to keep $H$ large as a safe choice. We present the hyperparameter values we used in our sweep in Table~\ref{tab:alpha_H}. The authors perform an ablation with DPPO for the square task in their appendix, and we received the hyperparameters from the authors for the same. After looking at their implementation, we observed that they train their RL policies with an expanded action space with an action horizon while we implement the RL policies in the original form with a single action.
\begin{table}[h]
    \vspace{1em}
    \centering
    \begin{tabular}{|c|c|c|c|}
         \hline
         \textbf{Environment} &\textbf{Base Policy} & $\boldsymbol\alpha$ & \textbf{H} \\
         \hline
         Lift & GMM & 0.1, 0.2, 0.05  & 400k, 600k  \\
         \hline
         Can & GMM & 0.05,0.1,0.2,0.5  & 400k, 600k, 800k \\
         \hline
         Square & GMM & 0.05, 0.1, 0.5 & 750k, 1M \\
         \hline
         Kitchen Complete & Diffusion & 0.1, 0.2, 0.3 & 400k, 600k  \\
         \hline
         Can & Diffusion & 0.2, 0.5 & 400k \\
         \hline
         Square & Diffusion & 0.05 & 500k \\
         \hline
         
    \end{tabular}
    \caption{Residual bound $\alpha$ and Progressive Exploration schedule $H$ for Policy Decorator.}
    \label{tab:alpha_H}
    \vspace{-1em}
\end{table}
}

\bibliographystyle{IEEEtran}
\bibliography{references}


 




\vfill

\end{document}